\documentstyle[twocolumn, art11]{article}

\advance \topmargin by -\headheight
\advance \topmargin by -\headsep
\evensidemargin \oddsidemargin
\begin{document}
\title{FOCUS ON ``ONLY" AND ``NOT"}

\author{Allan Ramsay\\
Dept. of Computer Science,
University College Dublin,
Belfield, Dublin 4, Ireland\\
{\tt allan@monkey.ucd.ie}}
\date{}
\maketitle
\def\medpara{\medskip\noindent}
\def\bigpara{\bigskip\noindent}

\def\implies{\rightarrow}
\def\endproof{\vrule width4pt height4pt}
\def\osb{[\hspace{-0.1em}[}
\def\csb{]\hspace{-0.1em}]}
\def\true{{\cal T}}
\def\false{\bot}
\def\and{\wedge}
\def\unify{\otimes}
\def\or{\vee}
\def\<{\langle}
\def\>{\rangle}
\def\unmodels{=\hskip -0.2em\mid}
\def\==>{\Longrightarrow}

\newcount\sentence

\sentence=1
\def\sent#1#2{
\edef #1{\the\sentence}
\begin{quote}
(\the\sentence)
{\em #2}
\end{quote}
\advance\sentence by1}

\begin{quote}
\noindent
Krifka [1993] has suggested that focus should be seen as a means of providing
material for a range of semantic and pragmatic functions to work on, rather
than as a specific semantic or pragmatic function itself. The current paper
describes an implementation of this general idea, and applies it to the
interpretation of {\em only} and {\em not}.
\end{quote}

\section{Background}

Consider the following sentences:

\sent{\SA} {I only borrowed your \underline{car}}

\sent{\SB} {I only borrowed \underline{your} car}

\sent{\SC} {I only \underline{borrowed} your car}

\medpara
All of them entail the same basic message, namely that I borrowed your car. In
addition to the basic message, however, they also carry information about
what I didn't do. (\SA) says that I didn't borrow any of your other
possessions,
(\SB) says that I didn't borrow anyone else's car, and (\SC) says that I didn't
do anything else to your car. It seems as though the word {\em only} and the
focus marker (indicated here by underlining the stressed word) combine to
add an extra message about what I didn't do.

A similar phenomenon appears to be taking place in the next set of sentences:

\sent{\SD} {I didn't steal your \underline{car}}

\sent{\SE} {I didn't steal \underline{your} car}

\sent{\SF} {I didn't \underline{steal} your car}

\medpara
Each of these says that I didn't steal your car, but again they each carry some
extra message. (\SD) says that I did steal something which belongs to you,
(\SE) says that I stole somebody's car, but not yours, and (\SF) says that I
did do something to your car (I probably borrowed it, though that is not
entailed by (\SF)).

Krifka [1993] argues that in (\SA--\SC) and (\SD--\SF), and in a number of
other
situations as well, the focus marker\footnote{\noindent
The term {\sc focus} has been used in a wide variety of ways. In the present
paper I simply use it to denote the part(s) of an utterance to which attention
is drawn by stress markers.}
is used to extract part of the
interpretation. Operators like {\em only} and so-called ``focussed negation"
then combine the extracted element of the interpretation with what was left
behind to carry complex messages of the kind discussed above.

The current paper shows how to implement this general notion, without following
Krifka's analysis in detail. The crucial point is the provision of some way of
storing the extracted part of the interpretation and making it available
when required. The interpretation of {\em only} and focussed negation is
fairly straightforward, so long as the treatment of the focussed item itself
is coherent.

\section {Abstraction and Focus}

The general aim of this paper
is to show how to use focus to decompose the interpretation
of a phrase into two parts, where one part is the interpretation of the
focussed item and the other is some object with which this can combine.
Suppose, for example, we thought that the VP {\em ate a peach} should be
interpreted as:

{\small
\begin{tabbing}
$\lambda A \exists Y ($\=$event(Y) \and type(Y, eat)$\\
\>$\ \and past(Y)
\and agent(Y, A)$\\
\>$\  \and \exists X (peach(X) \and object(Y, X)))$
\end{tabbing}}

\noindent
In other words, this VP is an abstraction over events where somebody ate a
peach.
Then we would want the two objects corresponding to the interpretation of
{\em ate a \underline{peach}} to be something like:

\begin{center}
{\small $\lambda Z peach(Z)$}
\end{center}

\noindent
and

{\small
\begin{tabbing}
$\lambda P \lambda A \exists Y ($\=$event(Y) \and type(Y, eat)$\\
\>$\ \and past(Y) \and agent(Y, A)$\\
\>$\ \and \exists X P.X \and object(Y, X)))$
\end{tabbing}}

\noindent
Here we have extracted the denotation of {\em peach} as the property of being
a peach, and converted the interpretation of the VP to an abstraction which
will combine appropriately with this property to reproduce the original
interpretation
\footnote{\noindent
You cannot freely mix $\lambda$-calculus and the truth functional connectives
of predicate calculus as we have here
without running into the paradoxes of self-reference --- Russell's paradox, the
Liar, and so on. The notation used in
this paper looks, for the sake of familiarity, like a combination of
$\lambda$-calculus and predicate calculus,
but is in fact grounded in the revision-based semantics of Turner's [1987]
{\sc property theory}.}.

Where else do we see a phenomenon of this kind? Consider the following phrases:

\sent {\SG} {the man who stole your bike}

\sent {\SH} {the man who I wanted you to meet}

\noindent
In (\SG) the property of being a man combines with the property of being
someone
who stole your bike to construct a contextually
minimal unique characterisation of the
relevant individual, and similarly in (\SH). To achieve this, we need to
interpret the relative pronouns in the two relative clauses
as leaving a hole
in the interpretation of clause and then abstracting with respect to that
hole. This is clear for (\SH), but it also holds for (\SG) if we want to
interpret a sentence like {\em a man stole a bike} as

{\small
\begin{tabbing}
$\exists Y ($\=$event(Y) \and type(Y, steal) \and past(Y)$\\
\>$\ \and \exists Z(man(Z) \and agent(Y, Z))$\\
\>$\ \and \exists X (bike(X) \and object(Y, X)))$
\end{tabbing}}

\noindent
where the quantifier introduced by the subject does not in fact have maximal
scope (an analysis I have argued for elsewhere [Ramsay 1992a]).

The treatment of (\SH) clearly requires much the same mechanism as we will
require if we want to deal with focus as outlined above, and this
may or may not also hold for (\SG). Any serious NLP system will include some
way of dealing with the interpretation of cases like (\SH), and almost any
such mechanism should be open to adaptation to deal with focus along the
suggested lines. One such approach is outlined below.

\section {Quantification, Presupposition, Abstraction and Focus}

We expect to interpret relative clauses (uncontroversially) and phrases with
focussed constituents (more controversially) as abstractions over the
interpretations of simple sentences. In order to construct interpretations of
the kinds of objects we are interested in, then, we have to start by looking
at simple sentences. The analyses presented in this paper start from the
following observations, most of which are fairly orthodox:

\begin{itemize}
\item
Indefinite NPs should be viewed as a way of introducing items (or preferably
sets of items) into the discourse. Universally quantified NPs say that all
items of the specified type satisfy some property.

\item
VPs should be viewed as a way of introducing events or possibly sets of events
into the discourse.

\item
If you construct interpretations by paraphrasing NL sentences into a formal
language which extends predicate calculus, you have to realise that the
scope of quantifiers in your paraphrases may not be determined by simple
structural properties of the source text.

\item
Definite NPs and other presuppositional constructions place constraints on the
discourse, so that a sentence containing the phrase {\em the man} will be
uninterpretable in contexts not containing a unique man (a version of
this point has been
made by, among others, Barwise \& Perry [1983], Kamp [1984],
Groenendijk \& Stokhof [1987]).

\item
There are interactions of scope between definite NPs and other types of
expression: in {\em Each man kills the thing he loves}, the presuppositional
construct {\em the thing he loves} requires the existence of a single
target of affection {\em per man}.
\end{itemize}

The standard way to deal with the potential discrepancy between where a
phrase appears and the width of its scope is by storing quantifiers on a
quantifier stack until the entire sentence has been interpreted, and then
using explicit information about the priority of various quantifiers to
sort things out [Cooper 1983, Vestre 1991]. The work reported here follows this
treatment, but extends it by introducing quantifier-like entities for
dealing with presuppositional items such as definite NPs (see Ramsay [1992b,
1994]
for a formal account of such {\sc constraints} on whether a sentence
is meaningful with respect to a situation). As an example, the sentence
{\em the woman stole a bike} is interpreted as

{\small
\begin{tabbing}
$\exists A\ $\=$A < now$\\
\>$ \and\ \iota B\ $\=$:($\=$\forall C \ member(C, B)  \implies\  woman(C)$\\
\>\>\>$\ \and\ |B|=1)$\\
\>\>$\exists D\ $\=$\forall E\ $\=$member(E, D)  \implies\  bike(E)$\\
\>\>\>$ \and\ |D|=1$\\
\>\>\>$ \and\ simple($\=$A, \lambda F\ $\=$event(F)$\\
\>\>\>\>\>$\  \and\  type(F, steal)$\\
\>\>\>\>\>$\  \and\ agent(F, B)$\\
\>\>\>\>\>$\  \and\ object(F, D))$
\end{tabbing}}

\noindent
This says that the relationship $simple$ holds between some past instant
$A$ and the property of being a certain sort of event. What sort of event?
One where a bike is stolen by someone (or rather, where a singleton set of
bikes is stolen). Writing something like $\iota B\ :(\forall C \ member(C,
B)  \implies\  woman(C) \and\ |B|=1) W$, where $W$ may contain
occurrences of $B$, says that $W$ holds for the contextually
unique individual $B$ which satisfies the restriction that $B$ is a woman
(is a singleton set of women). If this restriction fails to pick out a
unique individual the whole expression is meaningless in the context.

Most of this analysis is fairly orthodox. The two main points that might
require some defence are the analysis of aspect in terms of a relationship
between temporal objects and event types, which is discussed in [Ramsay 1993],
and the treatment of definite reference in terms of constraints on
meaningfulness. Neither of these is crucial to the remainder of the paper, but
if you don't like them you will have to replace them with something better,
and you are unlikely to find something which is both better and simpler.

The analysis above was obtained in a framework where quantifier scope is
determined on the basis of information explicitly associated with a form of
{\sc Cooper storage} ([Cooper 1983]), using
abstraction operators of the form
$\lambda W \exists X W$, $\lambda W \forall X W$ or
$\lambda W \iota X:R W$ which can be applied to a formula to bind its free
variables.
Within this framework, it is perfectly easy to deal with cases like (\SH) by
allowing the relative pronoun to add the expression $\lambda W \lambda X W$ to
the quantifier
store, annotated to specify that this expression has maximal scope.
If this expression is applied to a formula containing a free
occurrence of $X$ it will return an abstraction with respect to $X$ ---
exactly what we want. The requirement that this should have maximal scope will
ensure that $X$ is the last free variable in $W$.

But if we can use this mechanism to construct an abstraction as the
interpretation of a relative clause, we can also use it to construct an
abstraction as the interpretation of a phrase containing a focussed item. The
only extra work we have to perform is that we have to find somewhere to put
the interpretation of the focussed item itself. To do this, all that is
needed is
an extra feature {\tt focus} in the descriptions of linguistic items. The value
of {\tt focus} is the focussed item itself. {\tt focus} behaves like a GPSG
{\sc foot feature}, in that at most one daughter of an item can have a
non-vacuous value for {\tt focus}, and that if an item does have exactly one
daughter with a non-vacuous value for this feature then the item will share
this value with that daughter. {\tt focus} is thus very like the standard
feature {\tt slash} which is used for dealing with left extraposition ---
it is a foot feature whose value is some item which is somehow ``out of
position".

\section{Applications of Focus}

Once we have this mechanism, we can use it to construct interpretations of
sentences like (\SA)--(\SF). Consider, for instance, the example:

\sent {\SI} {I only borrowed a \underline{car}}

{\small\begin{tabbing}
$only($\=$\lambda A car(A),$\\
\>$\lambda B\ $\=$\exists C\ $\=$C < now$\\
\>\>\>$ \and\ \exists D\ $\=$\forall E\ $\=$member(E, D)  \implies\  B . E$\\
\>\>\>\>$\and\ |D|=1$\\
\>\>\>\>$ \and\ \iota F\ $\=$:($\=$\forall G member$\=$(G, F)$\\
\>\>\>\>\>\>\>$\ \implies$\=$\ speaker(G)$\\
\>\>\>\>\>\>$\ \and\ |F|=1)$\\
\>\>\>\>\>$simple($\=$C, K)))$
\end{tabbing}}

\noindent
where $K$ is
$\lambda H event(H) \and\  type(H, borrow)
\and\ agent(H, F)
\and\ object(H, D)$ (this has been extracted from the displayed formula to get
it inside the available space --- it is in fact part of that formula).

\noindent
This says that the relationship {\em only} holds between the property of
being a car and some other object. This is fine as far as it goes, but it
isn't worth very much unless we spell out the conditions under which
this relationship holds. The following meaning postulate does just that:

\begin{center}
{\small $\forall P \forall Q (only(P, Q) \implies\ Q.P
\and (\forall P' (Q.P' \implies P'=P)))$}
\end{center}

\noindent
In other words, if $only(P, Q)$ holds then $P$ satisfies $Q$ and nothing
else does. In the present case, the first of these consequences means that I
did indeed borrow a car:

{\small\begin{tabbing}
$$\=$\exists C\ $\=$C < now$\\
\>\>$ \and\ \exists D\ $\=$\forall E\ $\=$member(E, D)  \implies\  car(E)
\and\ |D|=1$\\
\>\>\>$ \and\ \iota F\ $\=$:($\=$\forall G \ member$\=$(G, F)
\implies\  speaker(G)$\\
\>\>\>\>\>$\ \and\ |F|=1)$\\
\>\>\>\>$simple($\=$C, K)))$
\end{tabbing}}

\noindent
where $K =
\lambda H event(H)  \and  type(H, borrow)
 \and agent(H, F) \and object(H, D)))$ has again been extracted to save space.
This was obtained from the meaning postulate by substituting $\lambda A
car(A))$ for $B$ and
using $(\lambda A car(A)).E \equiv car(E)$.

\noindent
The second consequence of the MP for $only$ says that there is no other
category of item which satisfies the abstraction --- that the only thing
I borrowed was a car.

If we put the focus somewhere else, we get another interpretation:

\sent {\SJ} {I only \underline{borrowed} a car}

{\small\begin{tabbing}
$only($\=$\lambda A\ $\=$\lambda B\ $\=$\lambda C\ $\=$event(C)  \and\  type(C,
borrow)$\\
\>\>\>\>$\ \and\ B . \lambda D(agent(C, D))$\\
\>\>\>\>$\ \and\ A . \lambda E(object(C, E)),$\\
\>$\lambda F\ $\=$\exists G\ $\=$G < now$\\
\>\>\>$ \and\ \exists H\ $\=$\forall I\ $\=$member$\=$(I, H)
\implies\  car(I)$\\
\>\>\>\>$ \and\ |H|=1$\\
\>\>\>\>$ \and\ \iota J\ $\=$:($\=$\forall K mem$\=$ber(K, J)$\\
\>\>\>\>\>\>\>$\ \implies\  speaker(K)$\\
\>\>\>\>\>\>$ \and\ |J|=1)$\\
\>\>\>\>\>$simple($\=$G, K))$
\end{tabbing}}

\noindent
where $K = (F . \lambda L(L . H)).\lambda M(M . J)))$

This says that {\em only} holds between a description of the type of event $C$
where somebody $B$ borrows something $A$,
and an abstraction over situations in which I did something to
some car.
Then the first consequence of {\em only} says that what I did to
this car was I borrowed it: substituting the description of the event type for
the abstracted variable $F$ produces $((\lambda A\lambda B\lambda C
event(C)$ $\and$ $type(C,$ $borrow)
$ $\and$ $B.\lambda D(agent(C,$ $D))
$ $\and$ $A.\lambda E(object(C,$ $E))).\lambda L(L . H)).\lambda M(M . J)$ as
the second argument of $simple$, and this reduces to
$\lambda C
event(C)$ $\and$ $type(C,$ $borrow)
$ $\and$ $agent(C,$ $J))
$ $\and$ $object(C,$ $H)$, which is what we want.
The second says that I didn't do anything else
to it.

Much the same will happen with

\sent {\SH} {I didn't \underline{steal} it}

{\small\begin{tabbing}
$not($\=$\lambda A\ $\=$\lambda B\ $\=$\lambda C\ $\=$event(C)  \and\  type(C,
steal)$\\
\>\>\>\>$\ \and\ B . \lambda D(agent(C, D))$\\
\>\>\>\>$\ \and\ A . \lambda E(object(C, E)),$\\
\>$\lambda F\ $\=$\exists G\ $\=$G < now$\\
\>\>\>$ \and\ \iota H\ $\=$:($\=$\forall I$\=$ \ member(I, H)
\implies\  neuter(I)$\\
\>\>\>\>\>$\ \and\ |H|=1)$\\
\>\>\>\>$\iota J\ $\=$:($\=$\forall K$\=$ \ member$\=$(K, J)$\\
\>\>\>\>\>\>\>\>$\ \implies\  speaker(K)$\\
\>\>\>\>\>\>$\ \and\ |J|=1)$\\
\>\>\>\>\>$simple($\=$G, K))$
\end{tabbing}}

\noindent
where $K = F . \lambda L(L . H).\lambda M(M . J)))$

Here we have a 2-place relation {\em not}, which is backed up by the following
MP:

\begin{center}
{\small$\forall P \forall Q (not(P, Q) \implies (\neg Q.P \and \exists
P'(Q.P')))$}
\end{center}

\noindent
This says that this form of negation holds between $P$ and $Q$ if $Q$ does
not hold of $P$, but does hold for some other entity $P'$. In the present
case, this means that I did do something to
it (whatever ``it" is), but what I did was not
stealing.

This contrasts with simple negation, with no focussed item, as in:

\sent {\SI} {I didn't steal it}

{\small\begin{tabbing}
$\neg($\=$\exists A\ $\=$A < now$\\
\>\>$ \and\ \iota B\ $\=$:($\=$\forall C \ member(C, B)  \implies\
neuter(C)$\\
\>\>\>\>$\ \and\ |B|=1)$\\
\>\>\>$\iota D\ $\=$:($\=$\forall E \ member$\=$(E, D)$\\
\>\>\>\>\>\>$\ \implies\  speaker(E)$\\
\>\>\>\>\>$\ \and\ |D|=1)$\\
\>\>\>\>$simple($\=$A, K)))$
\end{tabbing}}

\noindent
where $K = \lambda F event(F)  \and  type(F, steal) \and agent(F, D)
 \and object(F, B)))$

This simply says that it is not the case that there is a past
stealing event involving
me and it. The choice between the two is forced by the presence or absence
of a focussed item.

As a final example, consider a sentence which contains a focussed item but
no operator for using it up:

\sent {\SK} {A \underline{man} ate it}

\noindent
The analysis of this is an abstraction over kinds of individuals who ate it

{\small\begin{tabbing}
$\lambda A\ $\=$\exists B\ $\=$B < now$\\
\>\>$ \and\ \exists C\ $\=$\forall D\ $\=$member(D, C)  \implies\  A . D$\\
\>\>\>$\ \and\ |C|=1$\\
\>\>\>$ \and\ \iota E\ $\=$:($\=$\forall F \ mem$\=$ber(F, E)$\\
\>\>\>\>\>\>$\ \implies\  neuter(F)$\\
\>\>\>\>\>$\ \and\ |E|=1)$\\
\>\>\>\>$simple($\=$B, K)$
\end{tabbing}}

\noindent
with $K = \lambda G event(G) \and type(G, eat) \and agent(G, C) \and
object(G, E))$,
and
with the focus set as the description (including the semantic analysis) of the
focussed phrase {\em man}. This is just the kind of object required for a
discourse operator such as contrast or elaboration --- exactly which such
operator is appropriate depends on factors not visible in (\SK) itself, but
whatever it is it will require a pair of arguments of this kind.

\section{Conclusions}

The discussion above shows what can be achieved by treating focus as a
syntactic marker which makes information available to a variety of
operators. The mechanism for doing this involves introducing a foot feature
to carry the focussed item around, and constructing
appropriate abstractions by using the standard quantifier scoping
mechanism which is required for other phenomena anyway. Different NLP systems
will deal with the syntax
and semantics of phenomena such as left- and right-extraposition in
different ways. What I have argued is that almost any approach to these
phenomena can be adapted to deal with focus as well. The examples in Section
4 showed how you can combine these analyses of focus with a variety of
operators to convey a range of interpretations of the same sequence of words.
It is important to recall at this point that the
interpretation language being used here is
a highly intensional logic which permits quantification
over arbitrary kinds of individual, including quantification over
properties and propositions. I have argued elsewhere that such a language is
required for a wide variety of phenomena. The interpretation of focus is
just another example.

\section{Implementation}

All the analyses in this paper were produced, and $\lambda$-reduced
(and turned into \LaTeX!), by a
version of the system described in [Ramsay 1992a]. This consists of a highly
lexical grammar with a compositional semantics, parsed via a bi-directional
head-driven chart parser. I believe it is virtually impossible to do this
kind of work without embodying it in a working system. You simply cannot
explore the consequences of doing something one way rather than another,
or of combining an analysis of this with an analysis of that, unless
activities such as compositional construction and
subsequent $\lambda$-reduction of interpretations is done for you by machine.

\medpara
{\large \bf References}

\newif\ifpagenum \newif\ifloc \newif\ifnoref
\def\PAGENUM#1{\ifpagenum : #1\fi}
\def\ref#1#2#3#4{
    \loctrue\pagenumtrue
    \item [] #1 (#2#3): #4.}
    % \noindent #1 (#2#3): #4.}
\def\krf#1#2#3{\ref {#1} {#2} {{}} {#3}}

\def\book#1#2#3{{\em #1}: #2, #3}
\def\collection#1#2#3#4#5#6{#1, in {\em #2} (#3): #4, #5\PAGENUM{#6}}
\def\journal#1#2#3#4{#1, {\em #2} #3\PAGENUM{#4}}
\def\conference#1#2#3#4{#1, {\em #2}\ifloc, #3\fi\PAGENUM{#4}}
\def\ms#1#2{{\em #1}, #2}

\def\AI{Artificial Intelligence}
\def\CL{Computational Linguistics}
\def\KL{Kluwer Academic Publishers}
\def\UP{University Press}

\def\PROPSII{Properties, Types and Meaning II: Semantic Issues}
\def\PROPSA{G. Chierchia, B.H. Partee \& R. Turner}
\def\SNLP{
{\em Semantics of Natural Language}, eds. D. Davison \& G. Harman,
Reidel, Dordrecht: 1975}

\def\Barwise{
\krf {Barwise J. \& Perry J.} {1983}
{\book  {Situations and Attitudes}
        {Bradford Books}
        {Cambridge, MA}}}
\def\CooperA{
\krf {Cooper R.} {1983}
{\book
    {Quantification and Syntactic Theory}
    {Reidel}
    {Dordrecht}}}
\def\GroenendijkC{
\krf {Groenendijk J. \& Stokhof M.} {1991}
{\journal
    {Dynamic Predicate Logic}
    {Linguistics \& Philosophy}
    {14}
    {39--100}}}
\def\Kamp{
\krf {Kamp H.} {1984}
{\collection
    {A Theory of Truth and Semantic Representation}
    {Formal Methods in the Study of Language}
    {eds. J. Groenendijk, J. Janssen \& M. Stokhof}
    {Foris Publications}
    {Dordrecht}
    {277--322}}}
\def\KrifkaB{
\krf {Krifka M.} {1993}
{\pagenumfalse\journal
    {Focus, Presupposition and Dynamic Interpretation}
    {Journal of Semantics}
    {10}
    {0}}}
\def\RamsayH#1{
\ref {Ramsay A.M.} {1992} {#1}
{\journal
    {Presuppositions and WH-clauses}
    {Journal of Semantics} {9}
    {251--186}}}
\def\RamsayJ#1{
\ref {Ramsay A.M.} {1992} {#1}
{\conference
    {Bare Plural NPs and Habitual VPs}
    {COLING-92}
    {Nantes}
    {226-231}}}
\def\RamsayK#1{
\ref {Ramsay A.M.} {1993} {#1}
{\ms
    {Aspect and Aktionsart Without Coercion}
    {submitted to ACL-94}}}
\def\RamsayL#1{
\ref {Ramsay A.M.} {1994} {#1}
{\pagenumfalse\conference
    {Focus on ``Only" and ``Not"}
    {COLING-94}
    {Kyoto}
    {0}}}
\def\RamsayM#1{
\ref {Ramsay A.M.} {1994} {#1}
{\collection
    {Meanings as Constraints on Information States}
    {Constraints, Language and Computation}
    {eds. C.J. Rupp, M.A. Rosner \& R.L. Johnson}
    {Academic Press}
    {London}
    {249--276}}}
\def\TurnerA{
\krf {Turner R.} {1987}
{\journal
    {A Theory of Properties}
    {Journal of Symbolic Logic}
    {52(2)} {455--472}}}

\def\VestreA{
\krf {Vestre E.} {1991}
{\conference
    {An Algorithm for Generating Non-redundant Quantifier Scopings}
    {Fifth Conference of the European Chapter of the Assoc. for \CL}
    {Berlin}
    {251--256}}}

\begin{description}
\Barwise
\CooperA
\GroenendijkC
\Kamp
\KrifkaB
\RamsayJ{a}
\RamsayH{b}
\RamsayK{}
\RamsayM{}
\TurnerA
\VestreA
\end{description}

\end{document}